\def\year{2018}\relax
\documentclass[letterpaper]{article} 
\usepackage{aaai18}  
\usepackage{times}  
\usepackage{helvet}  
\usepackage{courier}  
\usepackage{url}  
\usepackage{graphicx}  
\usepackage{amsmath}
\usepackage{amssymb}
\usepackage{verbatim}
\usepackage{bbm}
\usepackage{soul}
\usepackage{color}
\usepackage{booktabs}
\usepackage{subfig}
\usepackage{tabularx}

\DeclareMathOperator{\LSTM}{Bi-LSTM}

\begin{document}
%
\title{Towards Automatic Learning of Procedures from Web Instructional Videos}
\author{Luowei Zhou\\
Robotics Institute\\
University of Michigan\\
{\tt\small luozhou@umich.edu}
\And
Chenliang Xu\\
Department of CS\\
University of Rochester\\
{\tt\small Chenliang.Xu@rochester.edu}
\And
Jason J. Corso\\
Department of EECS\\
University of Michigan\\
{\tt\small jjcorso@eecs.umich.edu}
}

\maketitle
\begin{abstract}
The potential for agents, whether embodied or software, to learn by observing other agents performing procedures involving objects and actions is rich. Current research on automatic procedure learning heavily relies on action labels or video subtitles, even during the evaluation phase, which makes them infeasible in real-world scenarios.
This leads to our question: 
can the human-consensus structure of a procedure be learned from a large set of long, unconstrained videos (e.g., instructional videos from YouTube) with only visual evidence?
To answer this question, we introduce the problem of \textit{procedure segmentation}---to segment a video procedure into category-independent \textit{procedure segments}. Given that no large-scale dataset is available for this problem, we collect a large-scale procedure segmentation dataset with procedure segments temporally localized and described; we use cooking videos and name the dataset YouCook2.
We propose a segment-level recurrent network for generating procedure segments by modeling the dependencies across segments. The generated segments can be used as pre-processing for other tasks, such as dense video captioning and event parsing.
We show in our experiments that the proposed model outperforms competitive baselines in procedure segmentation.
\end{abstract}

\section{Introduction}
\label{sec:intro}

Action understanding remains an intensely studied problem-space, e.g., action recognition \cite{donahue2015long,wang2016temporal}, action detection \cite{singh2016multi,yeung2016end,shou2016temporal} and action labeling \cite{kuehne2016weakly,huang2016connectionist,bojanowski2014weakly}. These works all emphasize instantaneous or short term actions, which clearly play a role in understanding short or structured videos~\cite{ChXiXuCVPR2014}. However, for long, unconstrained videos, such as user-uploaded instructional videos of complex tasks---\textit{preparing coffee} \cite{kuehne2014language}, \textit{changing tires}~\cite{alayrac2016learning}---learning the steps of accomplishing these tasks and their dependencies is essential, especially for agents' automatic acquisition of language or manipulation skills from video \cite{yu2013grounded,al2017natural,yu2015learning}.

\begin{figure*}[t]
\centering
       \includegraphics[width=2\columnwidth]{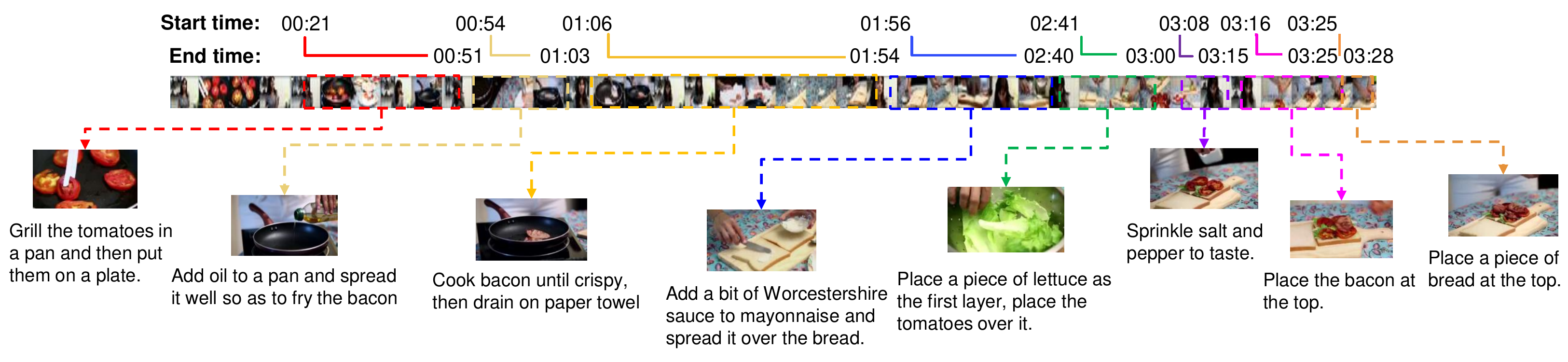}
       \caption{An example from the YouCook2 dataset on making a BLT sandwich. Each procedure step has time boundaries annotated and is described by an English sentence. Video from YouTube with ID: \url{4eWzsx1vAi8}.}
\label{fig:yc2}
\end{figure*}

We define \textit{procedure} as the sequence of necessary steps comprising such a complex task, and define each individual step as a \textit{procedure segment}, or simply \textit{segment} for convenience, inspired by~\cite{sener2015unsupervised,alayrac2016learning}.  
For example, there are 8 segments in the \textit{making a BLT sandwich} video shown in Fig.~\ref{fig:yc2}.  
We represent these segments by their start and end temporal boundaries in a given video.
Note that one procedure segment could contain multiple actions, but it should be conceptually compact, i.e., described with a single sentence. The number of procedure segments and their locations reflect human consensus on how the procedure is structured. 
Can this human-consensus procedure structure be learned by an agent?

To that end, we define the \textit{Procedure Segmentation} problem as: automatically segment a video containing a procedure into category-independent procedure segments. 
Although this is a new problem, there are two related, existing problems:  event proposal and procedure learning.
The event proposal problem~\cite{krishna2017dense} is to localize category-independent temporal events from unconstrained videos. Both event proposals and procedure segments can
contain multiple actions. 
However, the event proposal problem emphasizes the recall quality given a large amount of proposals, rather than the identification of a procedure (sequence of segments) from limited but necessary proposals. Events might overlap and are loosely-coupled but procedure segments barely overlap, are closely-coupled and usually have long-term dependencies.

The existing work in procedure learning is less-supervised than that of event proposals (no labels are given for the segments). It emphasizes video-subtitle alignment~\cite{malmaud2015s,bojanowski2015weakly} and discovery of common procedure steps of a specific process \cite{alayrac2016learning,sener2015unsupervised}. However, the methods proposed in these works make restrictive assumptions: they typically assume either language is concurrently available, e.g., from subtitles, or the number of procedure steps for a certain procedure is fixed, or both. Such assumptions are limited: extra textual input is unavailable in some scenarios; the subtitles or action sequences automatically generated by machines, e.g., YouTube's ASR system, are inaccurate and require manual intervention; and many procedures of a certain type, such a specific recipe, will vary the number of steps in different instances (process variation).

Unfortunately, work in neither of these two problems sheds sufficient light on understanding procedure segmentation, as posed above.  In this paper, we directly focus on procedure segmentation.  We propose a new dataset of sufficient size and complexity to facilitate investigating procedure segmentation, and we present an automatic procedure segmentation method, called \textit{ProcNets}.  

Our new dataset, called YouCook2\footnote{Dataset website: \url{http://youcook2.eecs.umich.edu}}, contains 2000 videos from 89 recipes with a total length of 176 hours. The procedure steps for each video are annotated with temporal boundaries and described post-hoc by a viewer/annotator with imperative English sentences (see Fig. \ref{fig:yc2}). To reflect the human consensus on how a procedure should be segmented, we annotate each video with two annotators, one for the major effort and the other one for the verification. To the best of our knowledge, this dataset is more than twice as large as the nearest in size and is the only one available to have both temporal boundary annotation and imperative English sentence annotation for the procedure segments. 

We then propose an end-to-end model, named \textit{Procedure Segmentation Networks} or \textit{ProcNets}, for procedure segmentation. 
\textit{ProcNets} make neither of the assumptions made by existing procedure learning methods: we do not rely on available subtitles and we do not rely on knowledge of the number of segments in the procedure. \textit{ProcNets} segments a long, unconstrained video into a sequence of category-independent procedure segments. ProcNets have three pieces: 1) context-aware frame-wise feature encoding; 2) procedure segment proposal for localizing segment candidates as start and end timestamps; 3) sequential prediction for learning the temporal structure among the candidates and generating the final proposals through a Recurrent Neural Network (RNN). The intuition is: when humans are segmenting a procedure, they first browse the video to have a general idea where are the salient segments, which is done by our proposal module. Then they finalize the segment boundaries based on the dependencies among the candidates, i.e., which happens after which, achieved by our sequential prediction module. 

For evaluation, we compare variants of our model with competitive baselines on standard metrics and the proposed methods demonstrate top performance against baselines. Furthermore, our detailed study suggests that ProcNets learn the structure of procedures as expected. 

Our contributions are three-fold. First, we introduce and are the first to tackle the category-independent procedure segmentation problem in untrimmed and unconstrained videos.
Second, we collect and distribute a large-scale dataset for procedure segmentation in instructional videos.
Third, we propose a segment-level recurrent model for proposing semantically meaningful video segments, in contrast to state-of-the-art methods that model temporal dependencies at the frame-level \cite{zhang2016video}. The output procedure segments of ProcNets can be applied for other tasks, including full agent-based procedure learning~\cite{yu2013grounded} or smaller-scale video description generation~\cite{yu2016video,krishna2017dense}.


\begin{figure*}[t]
\centering
   \includegraphics[width=1\linewidth]{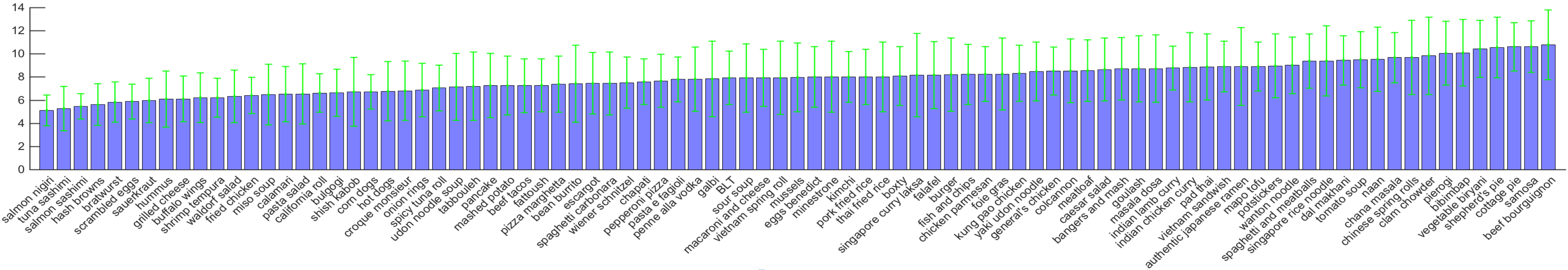}
   \caption{Mean and standard deviation of number of procedure segments for each recipe.}
\label{fig:clipnum}
\end{figure*}

\begin{figure*}
\def\tabularxcolumn#1{m{#1}}
\begin{tabularx}{\linewidth}{@{}cXX@{}}
\begin{tabular}{cc}
\subfloat[Distribution of video duration.]{\includegraphics[width=0.45\linewidth]{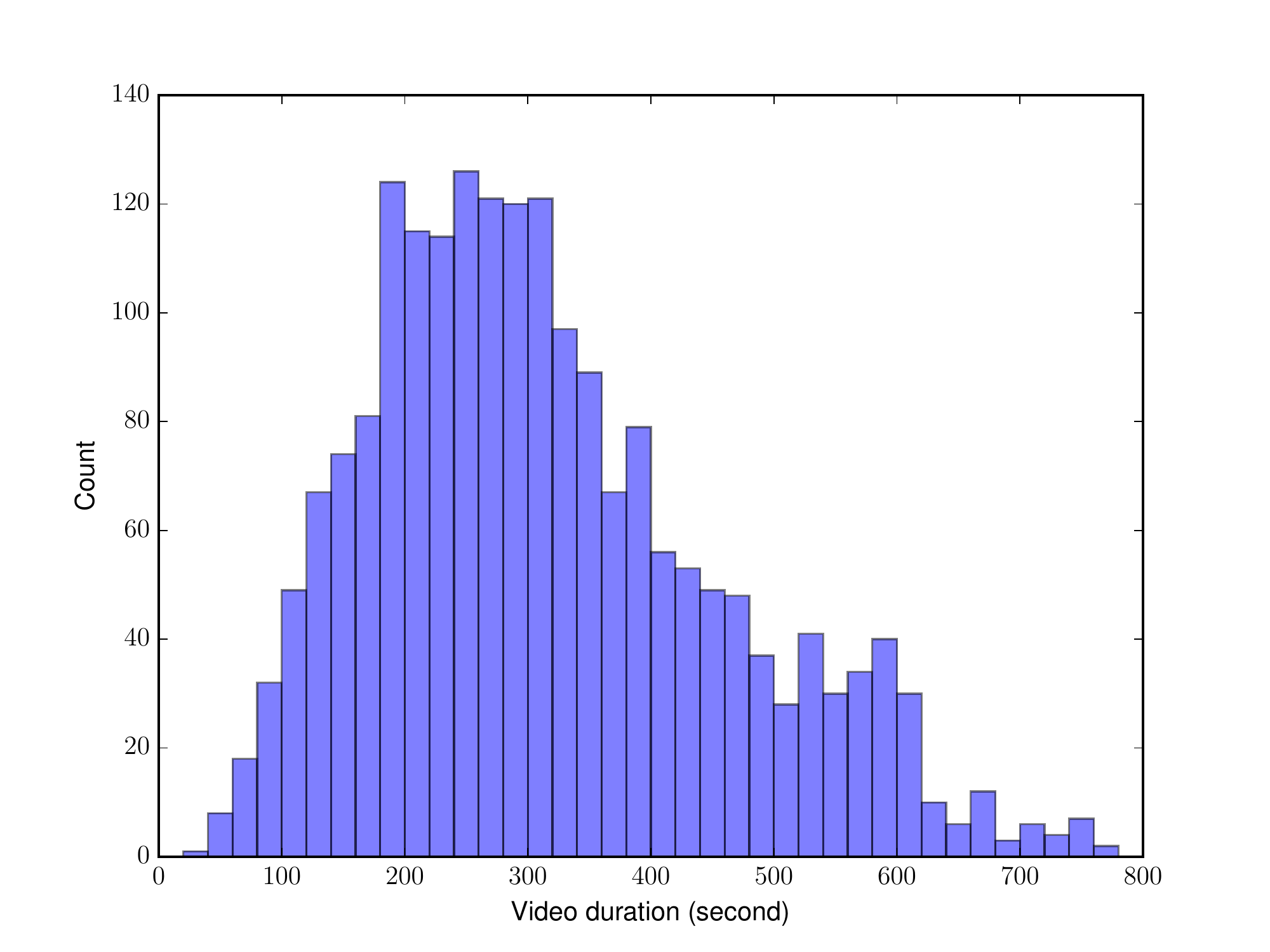}} &
\subfloat[Distribution of segment duration.]{\includegraphics[width=0.45\linewidth]{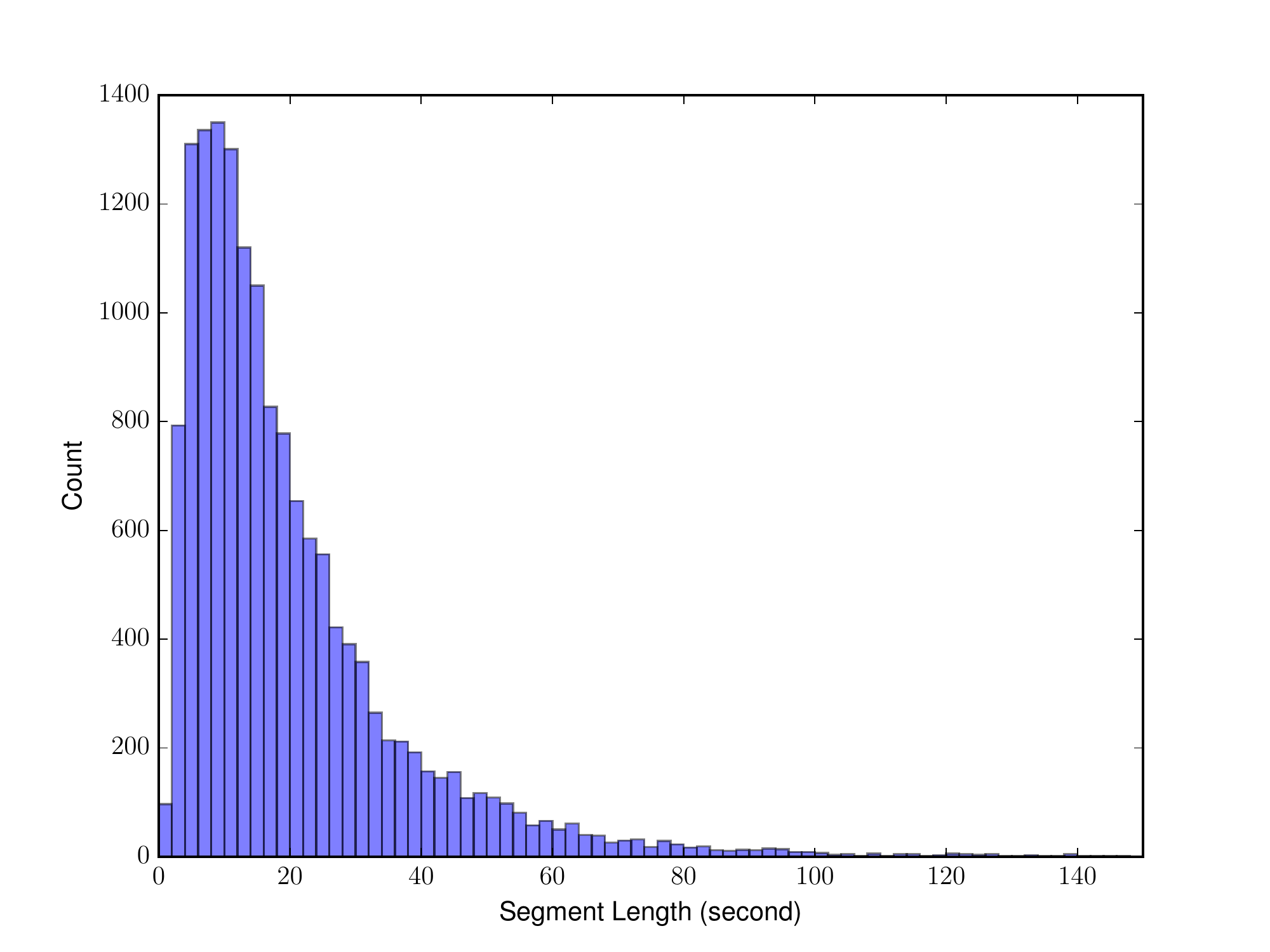}} \\
\end{tabular}
\end{tabularx}
\caption{YouCook2 dataset duration statistics.}\label{fig:stats}
\end{figure*}

\section{Related Work}

The approaches in action detection, especially the recent ones based on action proposal \cite{escorcia2016daps}, inspire our idea of segmenting video by proposing segment candidates. 
Early works on action detection mainly use sliding windows for proposing segments \cite{gaidon2013temporal,oneata2013action}. 
More recently, Shou et al. \cite{shou2016temporal} propose a multi-stage convolutional network called Segment CNN (SCNN) and achieves state-of-the-art performance \cite{THUMOS14}.
The most similar work to ours is Deep Action Proposals, also DAPs \cite{escorcia2016daps,krishna2017dense}, where the model predicts the likelihood of an action proposal to be an action while in our case segment proposal. DAPs determines fixed proposal locations by clustering over the ground-truth segments, while our model learns to localize procedures with anchor offsets, which is a generalization of the location pattern from training to testing instead of directly transferring.

Another topic similar to ours is action segmentation or labeling~\cite{kuehne2016end,kuehne2016weakly,huang2016connectionist,bojanowski2014weakly}. It addresses the problem of segmenting a long video into contiguous segments that correspond to a sequence of actions. Most recently, Huang et al. \cite{huang2016connectionist} propose to enforce action alignment through frame-wise visual similarities. Kuehne et al. \cite{kuehne2016weakly} apply Hidden Markov Models (HMM) to learn the likelihood of image features given hidden action states.
Both methods focus on the transitions between adjacent action states, leaving long-range dependencies not captured. Also, these methods generally assume contiguous action segments with limited or no background activities between segments. Yet, background activities are detrimental to the action localization accuracy \cite{huang2016connectionist}. We avoid these problems with a segment proposal module followed by a segment-level dependency learning module.

\section{YouCook2 dataset}
Existing datasets for analyzing instructional videos suffer from either limited videos~\cite{alayrac2016learning,sener2015unsupervised} or weak diversity in background~\cite{regneri2013grounding} and activities~\cite{kuehne2014language}). They provide limited or no annotations on procedure segment boundaries or descriptions~\cite{sigurdsson2016hollywood}. To this end, we collect a novel cooking video dataset, named YouCook2. 

YouCook2 contains 2000 videos that are nearly equal-distributed over 89 recipes. The recipes are from four major cuisine locales, e.g., Africa, Americas, Asia and Europe, and have a large variety of cooking styles, methods, ingredients and cookwares. The videos are collected from YouTube, where various challenges, e.g., fast camera motion, camera zooms, video defocus, and scene-type changes are present. Table~\ref{tbl:datasetcomp} shows the comparison between YouCook2 and other commonly-used instructional video datasets, e.g., YouCook \cite{das2013thousand}, MPII \cite{rohrbach2012database}, 50Salads \cite{stein2013combining}, Coffee \cite{alayrac2016learning}, Breakfast \cite{kuehne2014language} and Charades \cite{sigurdsson2016hollywood}.

Most of the datasets mentioned above have temporally localized action annotations. Compared to action segments, our procedure segments can contain richer semantic information and better capture the human-involved processes in instructional videos. Due to the variety of instructional processes and how each process can be performed, a fixed set of actions fails to describe the details in the video process (e.g., attributes and fine-grained objects). For example, the attribute ``crispy'' in the recipe step ``cook bacon until crispy then drain on paper towel'' (see Fig.~\ref{fig:yc2}) cannot be described by any action nor activity labels.

\begin{table}[t]
\centering
\caption{Comparisons of instructional video datasets. UnCons. stands for Unconstrained Scene and Proc. Ann. is short for Procedure Annotation. }
\label{tbl:datasetcomp}
    {\small
\begin{tabular}{cccc}
\toprule
\textbf{Name} & \textbf{Duration} &  \textbf{UnCons.} & \textbf{Proc. Ann.} \\
\midrule
YouCook & 140 m & Yes & No \\
MPII & 490 m & No &  No \\
50Salads & 320 m & No & No \\
Coffee & 120 m & Yes & No \\
Breakfast & 67 h & Yes & No \\
Charades & 82h & Yes & No \\
\midrule
\textbf{YouCook2} & \textbf{176h} & \textbf{Yes} & \textbf{Yes} \\
\bottomrule
\end{tabular}
    }
\end{table}

\begin{figure*}[t]
\centering
	\includegraphics[width=2\columnwidth]{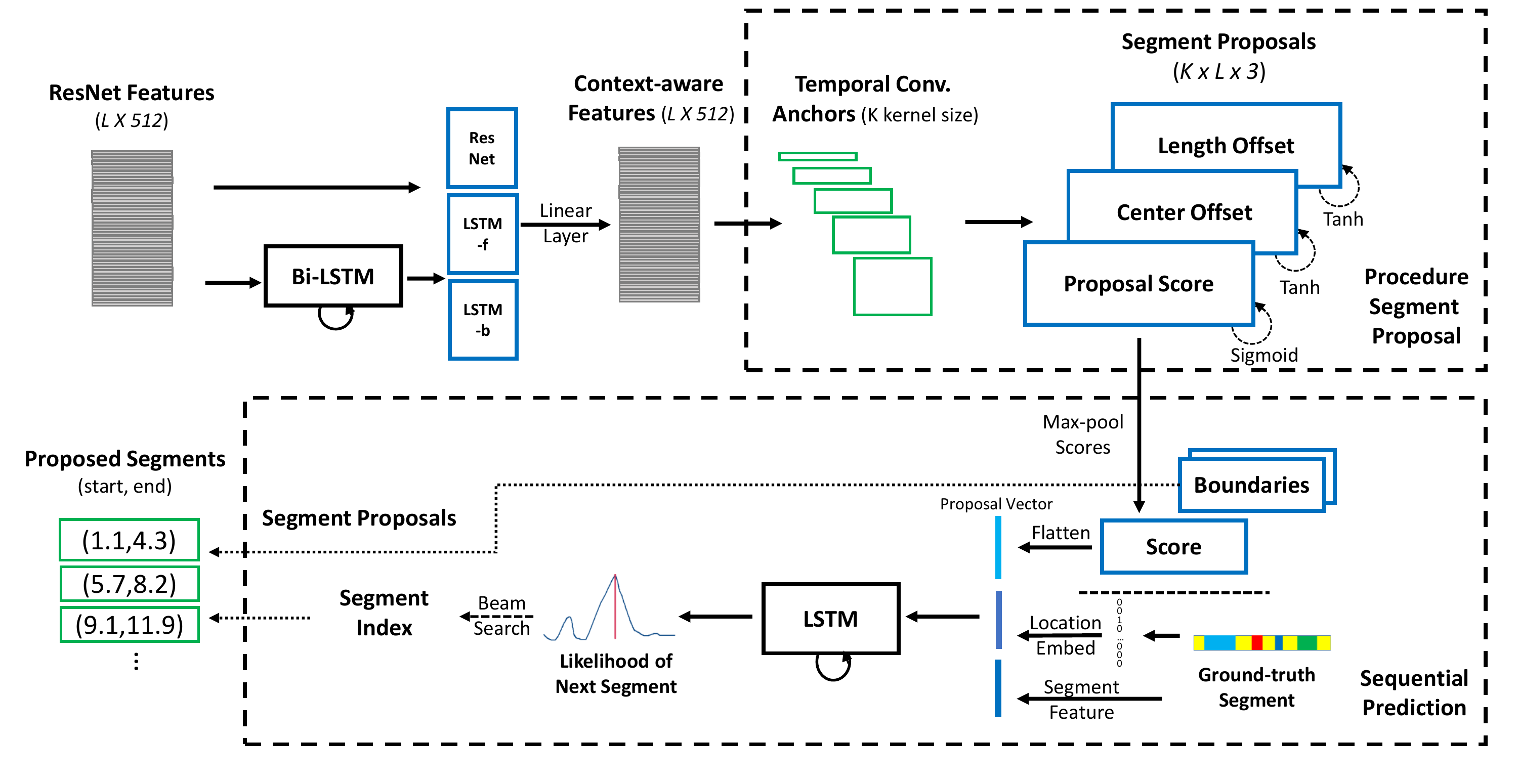}
	\caption{Schematic illustration of the ProcNets. The input are the frame-wise ResNet features (by row) for a video. The output are the proposed procedure segments. First, the bi-directional LSTM embeds the ResNet features into context-aware features. Then, the procedure segment proposal module generates segment candidates. Finally, the sequential prediction module selects the final proposals for output. During training, the ground-truth segments are embedded to composite the sequential prediction input, which are replaced with beam-searched segment in testing (as shown in the dashed arrows).}
\label{fig:tsl}
\end{figure*}

\subsection{Annotations}
Each video contains 3--16 procedure segments. The segments are temporally localized (timestamps) and described by English sentences in imperative form (e.g., \textit{grill the tomatoes in a pan}). An example is shown in Fig. \ref{fig:yc2}. The annotators have access to audio and subtitles but are required to organize and summarize the descriptions in their own way. As indicated in prior work \cite{baldassano2017discovering}, people generally agree with boundaries of salient events in video and hence we collect one annotation per video. To reflect the human consensus on how a procedure should be segmented, we annotate each video with two annotators, one for the major effort and the other one for verification. We also set up a series of restrictions on the annotation to enforce this consensus among different annotators. We have found that consensus is comparatively easy to achieve given the grounded nature of the instructional video domain.

Note that in this paper, we only use the temporal boundary annotations. We make the recipe descriptions available for future research.

\subsection{Statistics and Splits}
The average number of segments per video is 7.7 and the mean and standard deviation of the number of procedure segments per recipe are shown in Fig. \ref{fig:clipnum}.
The distribution of video duration is shown in Fig. \ref{fig:stats}(a). The total video length is 175.6 hours with an average duration of 5.27 min per video. All the videos remain untrimmed and can be up to 10 min.
The distribution of segment durations is shown in Fig. \ref{fig:stats}(b) with mean and standard deviation of 19.6s and 18.2s, respectively. The longest segment lasts 264s and the shortest one lasts 1s.
For the recipe descriptions, the total vocabulary is around 2600 words.

We randomly split the dataset to 67\%:23\%:10\% for training, validation and testing according to each recipe. Note that we also include unseen recipes from other datasets for analyzing the generalization ability of the models discussed. 

\section{Procedure Segmentation Networks}

We propose Procedure Segmentation Networks (ProcNets) for segmenting an untrimmed and unconstrained video into a sequence of procedure segments. 
We accomplish this by three core modules: 1) context-aware video encoding; 2) segment proposal module that localizes a handful of proposal candidates; 3) sequential prediction that predicts final segments based on segment-level dependencies among candidates.
At training, ProcNets are given ground-truth procedure segment boundaries for each video; no recipe categories or segment descriptions are given. At testing, for any given unseen video, ProcNets propose and localize procedure segments in the video based on their visual appearance and temporal relations.
The overall network structure is shown in Fig.~\ref{fig:tsl} and next, we explain each component.

\subsection{Context-Aware Video Encoding}

Define a video as $\mathbf{x} 
= \{x_1, x_2, \dots, x_L\}$, where $L$ denotes the number of sampled frames and $x_i$ is the frame-wise CNN feature vector with fixed encoding size. In this paper $L=500$ and encoding size is 512.
We use ResNet~\cite{he2015deep} as the appearance feature extractor for its state-of-the-art performance in image classification.
We then forward the ResNet features through a bi-directional long short-term memory ($\LSTM$)~\cite{graves2005framewise} as context encoding. 
The outputs (forward and backward) are concatenated with the ResNet feature at each frame and the feature dimension is reduced to the same as ResNet feature for a fair comparison. We call these frame-wise \textit{context-aware features}, denoted as $b_i = \LSTM(\mathbf{x})$.
Empirically, Bi-LSTM encoder outperforms context-free ResNet feature and LSTM-encoded feature by a relative 9\% on our evaluation metric.

\subsection{Procedure Segment Proposal}

Inspired by the anchor-offset mechanism for spatial object proposal, such as in Faster R-CNN~\cite{ren2015faster}, we design a set of $K$ explicit anchors for segment proposal. Each anchor has the length: $l_k \;(k=1,2,..,K)$ and their centers cover all the frames.

Each anchor-based proposal is represented by a proposal score and two offsets (center and length), from the output of a temporal convolution applied on the context-aware feature. The score indicates the likelihood for an anchor to be a procedure segment and the offsets are used to adjust the proposed segment boundaries. By zero-padding the video encoding at the boundaries (depending on anchor sizes), we obtain score and offset matrices of size $K \times L$ (see upper right of Fig. \ref{fig:tsl}) respectively, and hence the output of proposal module is $K\times L\times 3$. Sigmoid function and Tanh functions are applied for proposal score and offsets, respectively.

We formulate the proposal generation as a classification problem and proposal offset as a regression problem. The segment proposals are classified as procedure segment or non-procedure segment with binary cross-entropy loss applied. During training, the segment proposals having at least $0.8$ IoU (Intersection over Union) with any ground-truth segments are regarded as positive samples and these having IoU less than $0.2$ with all the ground-truth are treated as negative samples. We randomly pick $U$ samples from positive and negative separately for training. Then for the positive samples, we regress the proposed length and center offsets to the ground-truth ones from a relative scale. Given a ground-truth segment with center $c_g$ and length $l_g$, the target offsets ($\theta_c$, $\theta_l$) w.r.t. anchor (center $c_a$ and length $l_a$) are given by:
\begin{align}
\theta_c = \frac{c_g-c_a}{l_a} \;\;\;\;\; 
\theta_l = \log{\frac{l_g}{l_a}}
\label{eq:clipregress}
\enspace.
\end{align}
Smooth $l_1$-loss~\cite{ren2015faster} is applied in a standard way. For inference, the proposed offsets adjust the anchor location towards the final prediction location.

\subsection{Sequential Prediction}

\begin{figure}[t]
\centering
        \includegraphics[width=1\columnwidth]{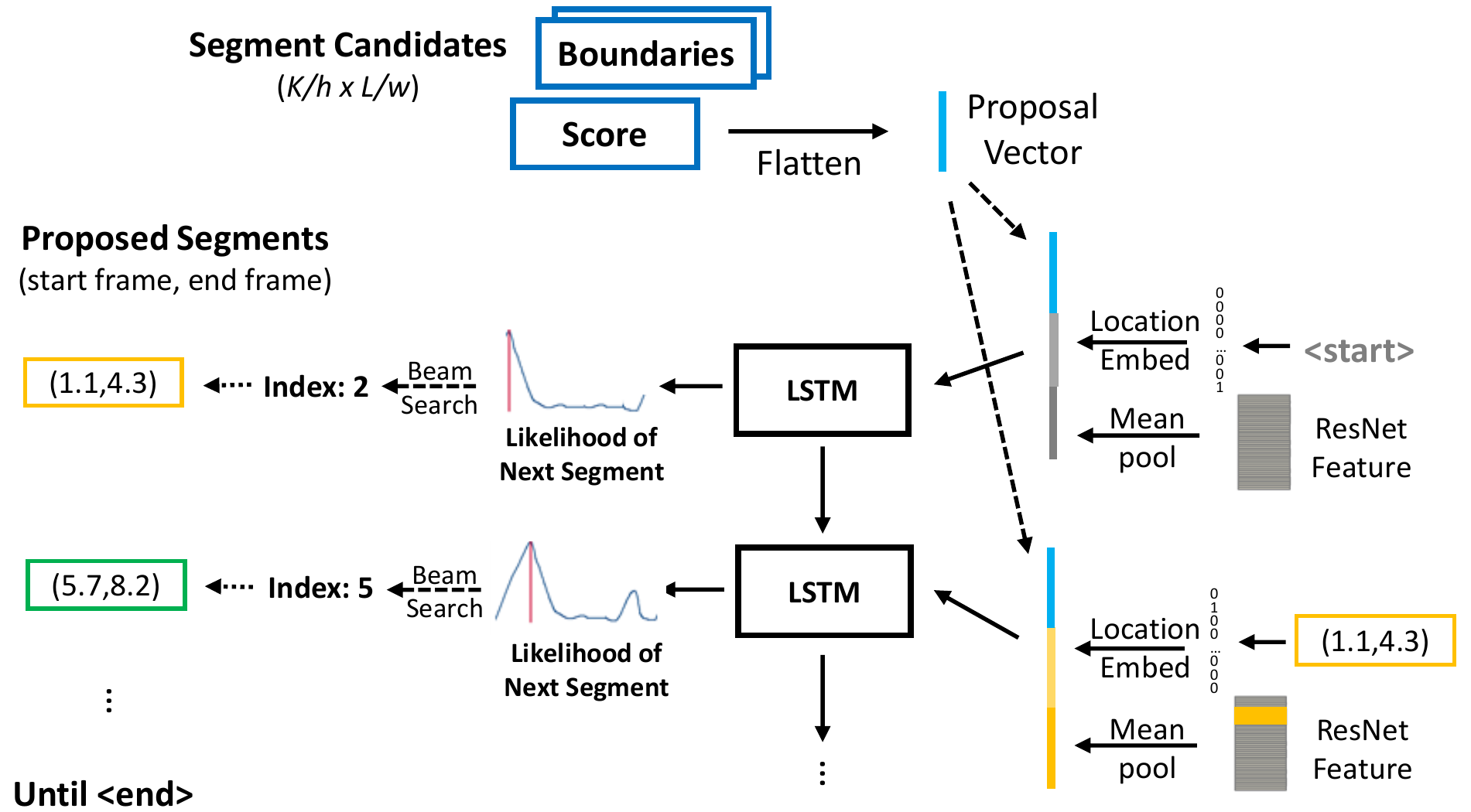}
   \caption{An example on sequential prediction during inference with unrolled LSTM. The $<$start$>$ token is feed into model at time 0. The previously generated segment is feed into model at time 1. Best view in color.}
\label{fig:seq_pred}
\end{figure}

Contrary to spatial objects, video procedure segments, by their nature, have strong temporal dependencies and yet ambiguous temporal boundaries. Therefore, we treat them differently. Recently, modeling frame-level temporal dependency in video has been explored~\cite{zhang2016video}. However, memorizing dependencies over enormous frames is still challenging for recurrent models to date \cite{singh2016multi}. In contrast, we propose to learn segment-level dependency because the number of proposal segments could be fewer so learning dependencies over segments are more tractable. By leveraging the segment-level dependency, we predict the sequence of procedure segments while dynamically determine the number of segments to propose.

We use long short-term memory (LSTM) for sequential prediction due to its state-of-the-art performance in sequence modeling \cite{xu2016msr,zhang2016video}. The input of LSTM is constructed from three parts: 1) Proposal Vector $\mathbf{S}$: max-pooled proposal scores from the proposal module, fixed over time; 2) Location Embedding $B_t$: a set of vectors that discretely encode the locations of ground-truth or previously generated segments; 3) Segment Content $C_t$: the visual features of the ground-truth or previously generated segments. The tuple $(\mathbf{S},B_t,C_t),\;t=1,2,...,N$, is concatenated as the input to LSTM at each time step $t$. Intuitively, when we learn to choose a few winners from a pool of candidates, we need to know who and how good they are (Proposal Vector), what they look like (Segment Content) and the target candidates (location embedding). We will detail each component after introducing the overall model first.

The softmax output of LSTM is the likelihood of each proposal being the next segment prediction. Therefore, the likelihood for the entire procedure segment sequence ${\epsilon_1,...,\epsilon_S}$ of a video can be formulated as:
\begin{align}
&\log p(\epsilon_1,...,\epsilon_S|\mathbf{S})\\
&=\sum_{t=1}^N \log p(\epsilon_t|\mathbf{S}, B_{t-1}, C_{t-1},\epsilon_0,...,\epsilon_{t-1}) \nonumber
\enspace,
\end{align}
where $\epsilon_0$ is the special $<$start$>$ segment token, $B_0$ is the embedding for the $<$start$>$ token, $C_0$ is the meal-pooled video feature over all frames, $B_{t-1}$ and $C_{t-1}$ are determined by $\epsilon_{t-1}$. The objective is to maximize the segment sequence likelihood for all training videos. We apply cross-entropy loss to the likelihood output $P_t$ at time step $t$ given the ground-truth segment index. During inference, we sample a sequence of segment indexes with beam search~\cite{vinyals2015show,donahue2015long}. In our experiments, simply set the beam size to 1 yields the best result, i.e., greedily picking the index with the maximal likelihood as the next proposed segment. The algorithm terminates when the special $<$end$>$ token is picked. An example is shown in Fig.~\ref{fig:seq_pred}. Next, we describe the three input vectors in details.

\noindent\textbf{Proposal Vector.}\quad As shown at the middle right of Fig.~\ref{fig:tsl}, we apply max-pooling to proposal score to filter out proposals with low proposal scores. The max-pooling kernel size is $h\times w$ and so as its stride, i.e., no overlapping. Empirically, $h=8$ and $w$ at 4 or 5 yields the best results. Given the filtered proposals (score and offsets), we flatten the proposal scores into a vector $\mathbf{S}$ by columns as Proposal Vector, which encodes the location and confidence information of all likely segment candidates in a video.


\begin{figure}[t]
\begin{center}
        \includegraphics[width=0.7\columnwidth]{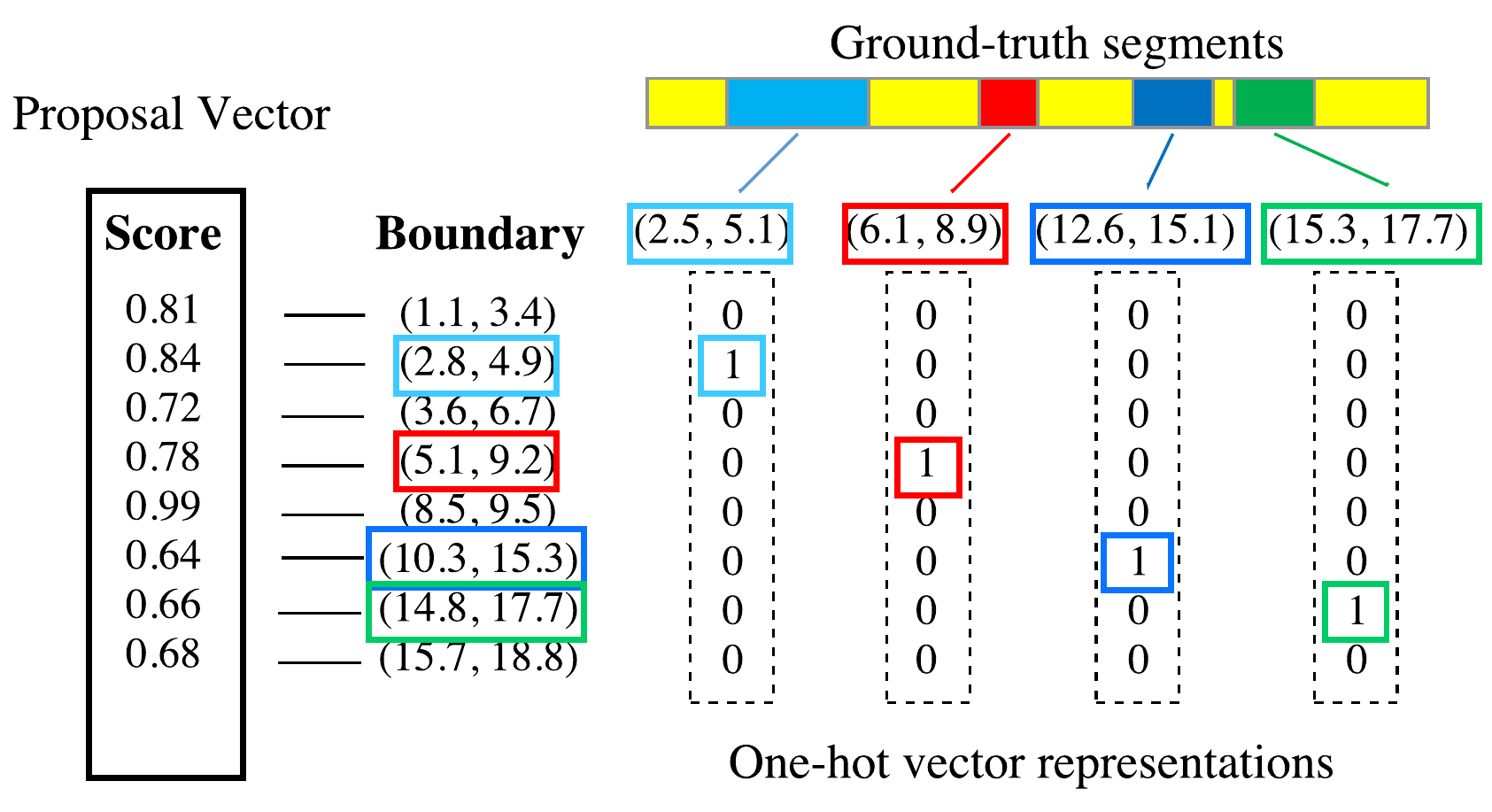}
\end{center}
   \caption{An example on converting ground-truth segments into one-hot vector representations from Proposal Vector.}
\label{fig:propvec}
\end{figure}

\noindent\textbf{Location Embedding.}\quad During training, each ground-truth segment is represented by a one-hot vector where the index of one matches to the nearest proposal candidate as illustrated in Fig.~\ref{fig:propvec}. This discrete representation of location is easier to learn than continuous location values. Through a trainable embedding matrix (similar to word embedding in language modeling), this one-hot vector maps to a vector, which we call Location Embedding vector and depicts the location information of a segment. This Location Embedding vector has the same size as the Proposal Vector. During testing, we greedily sample the softmax output of LSTM at previous time step to form location embedding for the current time step. Location Embedding represents the previous selected candidate, i.e., who we have and who we need next.

\noindent\textbf{Segment Content.}\quad We then encode the visual content for the candidate represented in the one-hot vector. We mean-pool the video ResNet feature bounded by the start and end timestamps of the candidate. Its dimension is reduced to the same as Proposal Vector by a fully-connected layer. Segment Content indicates what the candidate looks like.

\paragraph{Relations to Other Models.} To the best of our knowledge, we are the first to apply segment-level sequential modeling on category-independent procedure segments. The proposed model builds the video temporal structure without the need of knowing the hidden states such as in HMM. Note that there are other design choices. 

\noindent\textbf{Non-Maximal Suppression (NMS).}\quad In terms of proposal selection, a commonly adopted method in object detection~\cite{ren2015faster} or action detection~\cite{shou2016temporal} is NMS. This approach fails to capture the temporal structure or segment dependencies of instructional videos. We consider it as a baseline in our experiment along with our sequential prediction model.

\noindent\textbf{Other Time Sequence Models.}\quad Other methods for proposing segments have rigid model configurations, such as an HMM or pre-defined ``grammar'' for the whole video, which is infeasible for general video structure inference.

\subsection{Loss Function}
The loss function for procedure segmentation network consists of three parts, the binary cross-entropy loss for procedureness classification, the smooth $l1$-loss~\cite{ren2015faster} for offset regression and the cross-entropy loss for sequential prediction. The formulations are as follows: 
\begin{align}
&L= L_{cla}+\alpha_rL_{reg}+\alpha_sL_{seq} \\
&L_{cla}= -\frac{1}{U_p+U_n} (\sum_{i=1}^{U_p}\log(S_i^{(pos)})+\sum_{i=1}^{U_n}\log(1-S_i^{(neg)})) \nonumber \\
&L_{reg}= \frac{1}{U_p}\sum_{i=1}^{U_p}||B_i-B_i^{(gt)}||_{smooth-l1} \nonumber \\
&L_{seq} = -\frac{1}{N}\sum_{t=1}^{N}\log(P_t^T\mathbbm{1}_{t}^{(gt)}) \nonumber
\end{align}
where $U_p$ and $U_n$ are the number of positive and negative samples, respectively, $S_i^{(pos)}$ and $S_i^{(neg)}$represents their scores, $B_i^{(gt)}$ is the ground-truth boundary corresponding to positive sample $i$, $P_t$ is the softmax output of LSTM at time $t$ and $\mathbbm{1}_{t}^{(gt)}$ is one-hot vector of ground-truth segment index. Discount factors  $\alpha_r$ and $\alpha_s$ are applied to balance the contributions of the regression loss and sequential prediction loss, respectively. Empirically, equally weighting each part, i.e. $\alpha_r=\alpha_s=1$, yields good results.


\section{Experiments and Results}

In this section,  we benchmark our new dataset on procedure segmentation with competitive baselines and our proposed methods under standard metrics. We also show ablation studies, qualitative results and analysis on the procedure structure learned by our approach.

\noindent\textbf{Baselines.}\quad 
We compare our methods against state-of-the-art methods in video summarization and action proposal due to lack of direct baselines in our new problem. These methods include: 1) Video Summarization LSTM (vsLSTM) \cite{zhang2016video}, 2) Segment CNN for proposals (SCNN-prop) \cite{shou2016temporal}. 
The major difference between ProcNets and vsLSTM is, our model learns the segment-level temporal dependency while vsLSTM learns the frame-level temporal dependency. SCNN-prop is the proposal module of action detector SCNN, which achieves state-of-the-art performance in action proposal.\footnote{New results comparing DAPs and SCNN-prop: \url{https://github.com/escorciav/daps/wiki}} In addition, we also evaluate a uniform segment baseline (denoted as Uniform). 
Two variants of ProcNets are evaluated, one with all the modules (ProcNets-LSTM) and one that replaces sequential prediction with NMS (ProcNets-NMS). Finally, note that we compare with no action segmentation methods since these approaches require an action pool and directly model the finite action states (e.g., with HMM) which requires the ``grammar'' of the video procedure; both of these needs violate the core assumptions in this paper.

\noindent\textbf{Metrics.}\quad For procedure segmentation, we adopt two standard metrics for evaluating segment proposals: Jaccard \cite{bojanowski2014weakly} and mean Intersection over Union (mIoU). 
In Jaccard measure, the maximal intersection over prediction between all the final proposals and each ground-truth segment is computed and averaged. The individual Jaccard for each video is then averaged as the overall Jaccard. mIoU replaces the intersection over prediction in Jaccard with intersection over union (IoU). 
Hence, mIoU penalizes all the misalignment of segments while Jaccard only penalizes the partition of proposal beyond the ground truth. All the methods except for ProcNets-LSTM output 7 segments per video, determined by the average number of segments in the training set. Note that the average number of proposals from ProcNets-LSTM is also around 7, makes that a fair comparison.
Inspired by the average recall metric in action proposal~\cite{escorcia2016daps}, we also report the proposal averaged recall, precision and F1 score but with limited segments (10 per video), as motivated in Introduction section.

\noindent\textbf{Data Preprocessing.}\quad To preserve the overall information in the videos, we uniformly down-sample 500 frames for each video in YouCook2. The average sample rate is 1.58 fps.
To further enlarge the training samples, we temporally augment the data, i.e., sample each video 10 times with temporal shifts. Then, we extract the frame-wise ResNet-34 feature~\cite{he2015deep},\footnote{Torch implementation of ResNet by Facebook: \url{https://github.com/facebook/fb.resnet.torch}} pretrained on both ImageNet~\cite{deng2009imagenet} and 
MSCOCO caption~\cite{lin2014microsoft,zhou2016image}. Hence, each video is represented as a sequence of image spatial features. Local motion features are not used in our study; they may further improve performance.

\noindent\textbf{Implementation and Training Details.}\quad The sizes of the temporal conv. kernels (also anchor length) are from 3 to 123 with an interval of 8, which covers 95\% of the segment durations in training set. The 16 explicit anchors centered at each frame, i.e., stride for temporal conv. is 1. We randomly select $U=100$ samples from all the positive and negative samples respectively and feed in negative samples if positive ones are less than $U$.
Our implementation is in Torch. All the LSTMs have one layer and 512 hidden units. For hyper-parameters, the learning rate is $4\times 10^{-5}$. We use the Adam optimizer \cite{kingma2014adam} for updating weights with $\alpha=0.8$ and $\beta=0.999$. Note that we disable the CNN fine-tuning which heavily slows down the training process.

\subsection{Procedure Segmentation Results}
We report the procedure segmentation results on both validation and testing sets in Tab.~\ref{tbl:tc-results}. The proposed ProcNets-LSTM model outperforms all other methods by a huge margin in both Jaccard and mIoU. SCNN-prop \cite{shou2016temporal} suffers in our sequential segmentation task result from the lack of sequential modeling. vsLSTM \cite{zhang2016video} models frame-level temporal dependency and shows superior results than SCNN-prop. However, our model learns segment-level temporal dependency and yields better segmentation results, which shows its effectiveness. Uniform baseline shows competitive results and the possible reason is, in instruction videos, generally procedures span the whole video which favors segments that can cover the majority of video. For the rest experiments, all the results are on testing set.

\begin{table}[t]
\centering
\caption{Results on temporal segmentation. Top two scores are highlighted. See text for details.}
\label{tbl:tc-results}
    {\small 
\begin{tabular}{lcccc}
\toprule
 & \multicolumn{2}{c}{\textbf{validation}} & \multicolumn{2}{c}{\textbf{test}} \\
\textbf{Method (\%)} & \textbf{Jaccard} & \textbf{mIoU} &  \textbf{Jaccard} & \textbf{mIoU} \\
 \midrule
Uniform & 41.5 & \textbf{36.0} & 40.1 & \textbf{35.1} \\
vsLSTM & 47.2 & 33.9 & 45.2 & 32.2 \\
SCNN-prop & 46.3 & 28.0 & 45.6 & 26.7  \\
\midrule
ProcNets-NMS (ours) & \textbf{49.8} & 35.2 & \textbf{47.6} & 33.9  \\
ProcNets-LSTM (ours) & \textbf{51.5} & \textbf{37.5} & \textbf{50.6} & \textbf{37.0}  \\
\bottomrule
\end{tabular}
    }
\end{table}


\begin{table}[t]
\centering
\caption{Ablation study on LSTM input. We remove either Proposal Vector (as \textit{-Proposal Vec}), Location Embedding (as \textit{-Location Emb}) or Segment Content (as \textit{-Segment Feat}).}
\label{tbl:abla}
    {\small 
\begin{tabular}{lcc}
\toprule
 & \textbf{Jaccard} & \textbf{mIoU}\\
 \midrule
Full model & 50.6 & 37.0 \\
\;\;\textit{-Proposal Vec} & 47.6 & 36.1 \\
\;\;\textit{-Location Emb} & 46.2 & 35.1 \\
\;\;\textit{-Segment Feat} & 49.0 & 36.4 \\
\bottomrule
\end{tabular}
    }
\end{table}

\begin{table}[t]
\centering
\caption{Results on segment localization accuracy. Top two scores are highlighted.}
\label{tbl:det-accu}
    {\small 
\begin{tabular}{lccc}
\toprule
\textbf{Method (\%)} & \textbf{Recall} & \textbf{Precision} & \textbf{F1} \\
 \midrule
vsLSTM  & 22.1 & \textbf{24.1} & 23.0 \\
SCNN-prop  & \textbf{28.2} & 23.2 & \textbf{25.4} \\
\midrule
ProcNets-NMS & \textbf{37.1} & \textbf{30.4} & \textbf{33.4} \\
\bottomrule
\end{tabular}
    }
\end{table}

\noindent\textbf{Ablation study on sequential prediction.}\quad The input of the sequence modeling LSTM is the concatenation of three parts: Proposal Vector, Location Embedding and Segment Content. We remove either one of them as the ablation study. Results are shown in Tab.~\ref{tbl:abla}. Unsurprisingly, the proposal scores (Proposal Vector) play a significant role in determining the final proposals. When this information is unavailable, the overall performance drops by 6\% on Jaccard relatively.
The Location Embedding encodes the location information for ground-truth segments and is the most important component for procedure structure learning. Jaccard and mIoU scores drop by 8.7\% and 5.1\% relatively when location embedding is not available.
The segment visual feature has less impact on the sequence prediction, which implies the visual information represented in the video appearance feature is noisy and less informative.

\begin{figure*}[t]
\centering
\includegraphics[width=0.9\linewidth]{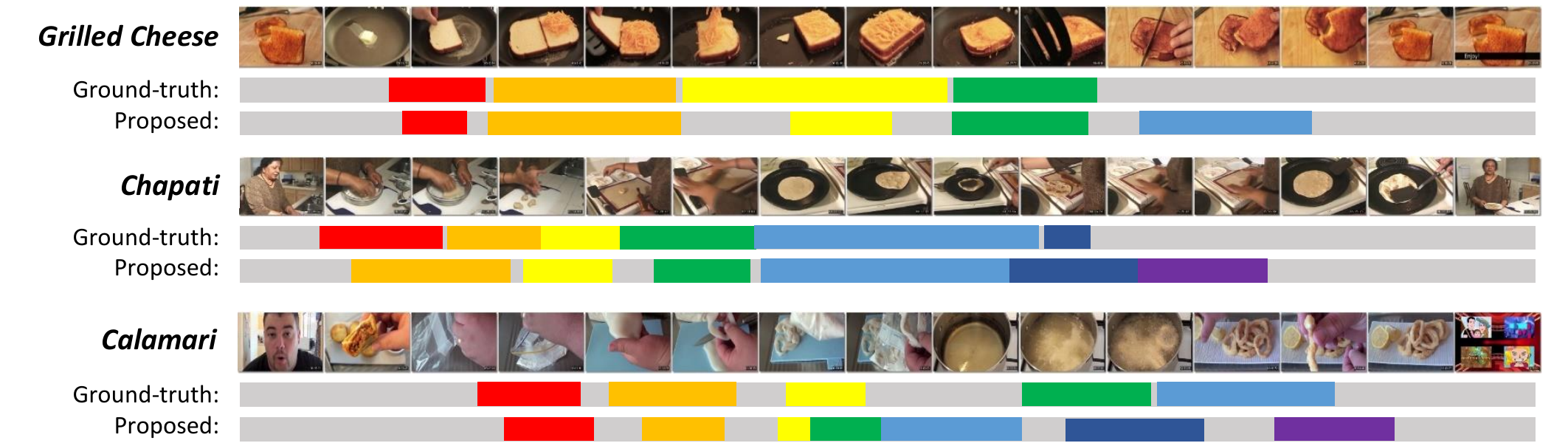}
\caption{Qualitative results from test set. YouTube IDs: \url{BlTCkNkfmRY}, \url{jD4o_Lmy6bU} and \url{jrwHN188H2I}.}
\label{fig:vis}
\end{figure*}

\begin{figure}[th!]
\centering
\includegraphics[width=1\linewidth]{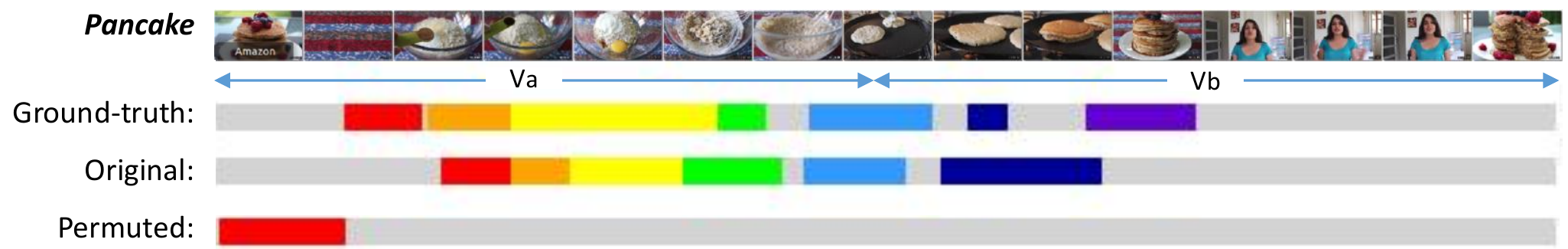}
\caption{An example output of ProcNets on the original and the permutated video. YouTube ID: \url{ejq2ZsHgwFk}.}
\label{fig:ba}
\end{figure}

\noindent\textbf{Proposal localization accuracy.}\quad We study the proposal localization problem when each model proposes 10 segments per video. 
Note that the metrics used here are not suitable for ProcNets-LSTM as they impose a fixed number of segments, where ProcNets-LSTM learns that automatically; nonetheless, we evaluate ProcNets-NMS for the quality of procedure segment proposal. 
%
The average recall, precision and F1 are shown in Tab. \ref{tbl:det-accu}. The IoU threshold for true positive is 0.5. SCNN-prop shows competitive localization results as expected. vsLSTM yields inferior localization accuracy even though it performs better than SCNN-prop on segmentation. Our proposed model has more than 9\% and 7\% higher recall and precision than the baselines.

\noindent\textbf{Qualitative results.}\quad We demonstrate qualitative results with videos from YouCook2 test set (see Fig.~\ref{fig:vis}). The model can accurately localize some of the segments and predict their lengths. Moreover, the number of segments proposed is adapted to individual videos and the model learns to propose fewer segments at the beginning and the end of the video, where usually no cooking processes happen. In the example of \textit{making Grilled Cheese}, ProcNets propose the fifth segment to cover the process of cutting bread slices into two pieces. This trivial segment is not annotated but is still semantically meaningful.

\noindent\textbf{Analysis on temporal structure learning.}\quad We conduct additional experiments to evaluate the temporal structure learning capability of ProcNets. For a given testing video, denote the first half as $V_a$ and the second half as $V_b$. We inverse the order of $V_aV_b$ to $V_bV_a$ to construct the permutated video. We evaluate our model on both original test set and the permutated test set. The performance of pre-trained ProcNets decreases by over a half in the permutated set and 10\%-20\% of the videos only have segments predicted at the beginning of $V_b$ (see Fig. \ref{fig:ba}). We believe reasons are two. 
First, the model captures the ending content in $V_b$ and terminates the segment generation within $V_b$. Second, the temporal structure of $V_a$ has no dependencies on $V_b$ and hence is ignored by the model.

\section{Conclusion}

We introduce a new problem called procedure segmentation to study human consensus on how a procedure is structured from unconstrained videos. Our proposed ProcNets take frame-wise video features as the input and predict procedure segments exist in the video. We evaluate the model against competitive baselines on the newly collected large-scale cooking video dataset with standard metrics and show significant improvements. Besides, ProcNets are capable of inferring the video structure by video content and modeling the temporal dependencies among procedure segments. For future work, there are two extensions of the current work. The first one is dense video captioning. The other one is weakly supervised segmentation, which is to first align the weak audio/subtitle signal with the video and then train our model with the aligned annotation.

\paragraph{Acknowledgement.}\quad This work has been supported in part
by Google, ARO W911NF-15-1-0354, NSF CNS 1463102, NSF CNS 1628987, NSF NRI 1522904 and NSF BIGDATA 1741472. This article solely reflects the opinions and conclusions of its authors and neither Google, ARO nor NSF. We sincerely thank our colleagues Vikas Dhiman, Madan Ravi Ganesh and Ryan Szeto for their helpful discussions. We also thank Mingyang Zhou, Yichen Yao, Haonan Chen, Carolyn Busch and Dali Zhang for their efforts in constructing the dataset.


\begin{small}
\bibliographystyle{aaai}
\bibliography{procnets}
\end{small}

\end{document}